# Objective Function Designing Led by User Preferences Acquisition

Patrick Taillandier and Julien Gaffuri

*Abstract*—Many real world problems can be defined as optimisation problems in which the aim is to maximise an objective function. The quality of obtained solution is directly linked to the pertinence of the used objective function. However, designing such function, which has to translate the user needs, is usually fastidious. In this paper, a method to help user objective functions designing is proposed. Our approach, which is highly interactive, is based on man-machine dialogue and more particularly on the comparison of problem instance solutions by the user. We propose an experiment in the domain of cartographic generalisation that shows promising results.

## I. INTRODUCTION

Artificial systems are more effective than humans to solve many problems. One of the reasons is their computing capacity that allows them to tests many possibilities in a short period of time. However, in order to get good results, an artificial system has to know what it is searching, i.e. what type of solutions is expected. Unfortunately, while human experts can easily give a qualitative evaluation of the quality of a problem solution or order several solutions in terms of quality, it is often far more difficult for them to express their expectations in a formal way that can be used by artificial systems. This problem is particularly complex when numerous measures are used to characterise a solution and when no simple links can be found between these measures values and the solution quality.

This paper deals with the problem of the formalisation of the user outcome expectation from the system, into a form usable by artificial systems. In this context, we propose an approach aiming at translating the user needs into an objective function thanks to dialogue between the user and the system.

In Section 2, the general context of our work is introduced. Section 3 is devoted to the presentation of our approach. Section 4 describes an application of our approach to cartographic generalisation. We present a real case study that we carried out as well as its results. Section 5 concludes and presents the perspectives of this work.

## II. CONTEXT

### A. Optimisation problem and objective function

Many real world problems can be expressed as optimisation problems. Solving such a problem consists in finding, among all possible solutions of the problems, the one that maximises an objective function. This function characterises the quality of a solution. Its definition is a key point of the resolution of optimisation problems [19]-[20]. Indeed, the goal of the resolution of an optimisation problem is to find the solution that maximises (or minimises) this function. Thus, if the objective function is not in adequacy with the real quality of a solution, the solutions that will be found will never be good. Many works were interested in the definition of such function for specific problems [14]-[21] but few proposed general approach for helping the user of an optimisation system to define it.

### B. Related Works

The problem of objective function definition and more generally of user need definition is a complex problem which was studied in various fields.

A first approach to solve this problem is to use supervised machine learning techniques. These techniques consist in inducing a general model from examples labeled by a user/expert. In this context, it is possible to learn an objective function from examples evaluated by a user. This approach was used in several works. For example, [21] used this approach in the domain of computer vision, [6], in the learning of cognitive radio.

A second approach consists in establishing a man-machine dialogue to converge toward a formalisation of the user needs. Reference [5] proposes to use such approach in order to help users to create original map legends. This work proposed to use map samples to establish a dialogue between the user and the system. This dialogue allows the system to retrieve the user preferences, and thus to design a suitable legend that respects the user expectations as well as cartographic constraints (to ensure the map readability). In the same way, [13] proposes to use map samples to capture user needs in terms of geographic information. Our work is taking place in the continuity of these two works. We propose to use the same approach based on a dialogue between the user and the system established through the presentation of samples.

P. Taillandier is with the UMI UMMISCO of IRD, 209, 32 avenue Henri Varagnat,93143 Bondy, France and with the UMI 209 of IFI, MSI, ngo 42 Ta Quang Buu, Ha Noi, Viet Nam (e-mail: patrick.taillandier@gmail.com).

J. Gaffuri is with the COGIT lab of IGN , 73 avenue De Paris, 94165 Saint-Mandé, France (e-mail: julien.gaffuri@ign.fr).

## III. PROPOSED APPROACH

### A. General approach

As stated in the introduction, if experts often have difficulties to express in a formal way their needs from a system, it is far easier for them to compare different solutions of a problem and to point out their preferences. Thus, we propose to base our user need definition approach on the presentation of comparisons between solutions to the user. Each comparison is composed of two solutions for a same problem instance. The user can give his preferences toward these two solutions to the system, i.e. the solution that he prefers if there is one. The system then automatically build the evaluation function from the whole set of preferences.

Our general approach, presented Figure 1, is composed of 3 steps: the first one consists in generating a set of pairs of solutions to compare (called "comparisons set"); the second one consists in capturing the user preferences by asking the user to select its favourite solution for each comparison; the last step consists in using these captured preferences to automatically build the objective function that will represent the user expectations toward the optimisation system.

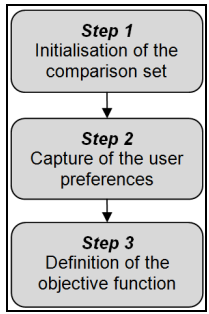

Fig. 1. General approach

In the following sections, we described each of these three steps.

### B. Initialisation of the comparison set

The first step of our approach consists in generating a set of comparisons that will be used to capture the user needs. We defined a comparison as a set of two solutions for a same problem instance.

The generation of the comparison set depends on the context of use of our approach. For example, in the case where a set of instances of the considered problem is available and where this set is too big to take into account all available instances, a sampling method has to be used in order to select a subset of problem instances. The subset has to be representative of the whole set in order to capture more pertinently the user preferences in a generic objective function.

Each selected instance has to be solved in order to obtain at least two solutions for it. For each couple of solutions, a comparison is created and is added to the comparison set.

### C. Capture of the user preferences

The second step of our approach consists in capturing the user preferences. Figure 2 presents our approach: at each iteration, a comparison is selected between all available ones (the comparison set). Then, the user defines its preferences, i.e. between the two solutions, the one that he finds better. The user can also define that the two solutions are as good or as bad. This sequence is reiterated until an ending criterion is checked. An example of ending criterion can be to stop the cycle when a specific number of comparisons have been presented to the user.

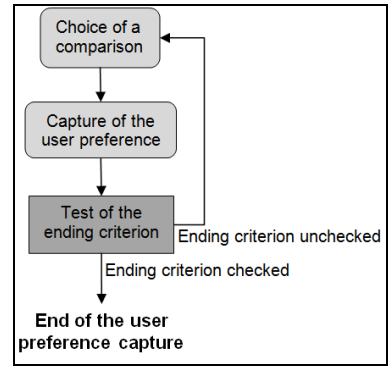

Fig. 2. User preference capture approach

The main question of this step concerns the choice of a comparison to propose to the user at each iteration. How to choose that comparison? To guide this choice, we propose to use *comparison choice strategies*: a *comparison choice strategy* allows the choice of the next comparison among a set of comparisons according to a specific strategy.

In this paper, we propose four different *comparison choice strategies*:

- *Measure consistency analysis*: this strategy consists in choosing a comparison where the two solutions are equivalent in terms of measure values. The goal is to analyse the consistency of the measure set. If the user prefers one of the two solutions whereas they are equivalent in terms of measure values, it means that the measure set is not pertinent and does not allow to well-characterised the solution quality.
- *Measure evolution analysis*: this strategy consists in choosing a comparison where the value of only one measure changes between the two solutions. The goal is to analyse how the quality of a solution evolves according to the evolution of the value of this measure.
- *Order of preference between two measures*: this strategy consists in choosing a comparison where the values of only two measures change between the two solutions. The goal is to compare the relative importance of each measure for the computation of the solution quality.
- *Random comparison*: this strategy consists in choosing randomly a comparison in the comparison set.

In order to define a global strategy of user preference capture,

we propose to chain different *comparison choice strategies*. Indeed, in a first step, we propose to apply the *measure consistency analysis* strategy in order to check the pertinence of the measure set. If this one is not pertinent, the objective function obtained at the end of the user need definition process will certainly not be perfect. Then, in a second step, we propose to apply *the measure evolution analysis* strategy for each measure. This step allows a better understanding of the link existing between the evolution of a measure and the evolution of the solution quality. The third step consists in applying the *order of preference between two measures* strategy to compare by couple the relative importance of each measure. In the last step, we propose to apply he *random comparison* strategy.

### D. Definition of the objective function

The last step of our approach consists in designing an objective function from the user preferences.

We propose to formulate the objective function as a set of regression rules. Each regression rule is associated to a weighted means. The interest of such representation is to be easily interpretable by domain experts and thus to facilitate the objective function validation.

Let $M$ be the set of measures, $w_i$ the weight associated to the measure $i$ and $Val_i(sol)$, the value of the measure $i$ for the solution $sol$ belonging to the whole possible solution set SOL.

We define the measures of $M$ such as:
$$\forall sol \in SOL, \forall i \in M, VAL\_MIN \leq val_i(sol) \leq VAL\_MAX$$

with VAL_MIN and VAL_MAX real.

Each regression rule is defined as follows:

if condition then $quality(sol) = \dfrac{1}{\sum_{i \in M} w_i} \times \sum_{i \in M} w_i \times Val_i(sol)$

An Example of objective function is presented in Section IV.E.3.

Building an objective function consists in finding a set of regression rules (with, for each of them, a condition and the weight values) from the preferences given by the users on the samples. As presented Figure 3, to solve this problem, we propose to use an approach based on the search of the best weights and eventually on the partitioning of the measures set (which correspond to the addition of new regression rules).

At the initial stage, the objective function is composed of only one regression rule, such as the measure space is composed of only one partition. At the first step, the system searches a weight assignment that maximises the adequacy between the objective function and the user preferences. If this weight assignment is in total adequacy with the user preferences, the process ends; the objective function is composed of only one regression rule. Otherwise, new regression rules are introduced: the system computes partitions of the measure set in order to detect the parts of the measure set that are not compatible with the others. Then, a new weight assignment is searched again for all regression rules, by considering all partitions built at the same time. If the weight assignment obtained after the partitioning allows to get a better result than the previous one, it is kept; otherwise, the system backtracks to the previous objective function and end the evaluation function building process. This partitioning procedure is recursively repeated until the learnt objective function allows to obtain the given user preferences or until no more improvement of the objective function is obtained.

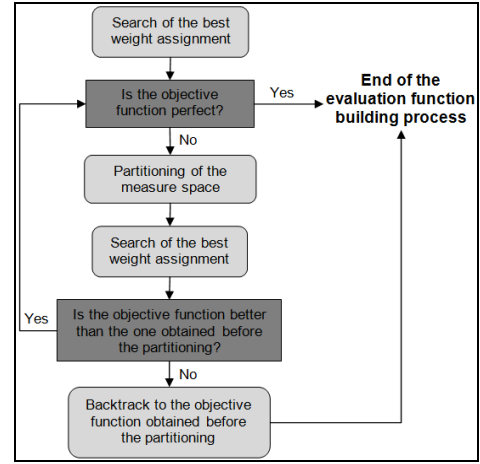

Fig. 3. Approach of evaluation function building

#### 1) Search of the best weight assignment

We propose to formulate the problem of best weight assignment as a minimisation problem. We define a global error function that represents the inadequacy between the evaluation function (and thus the weight assignment) and the user preferences. The goal of the best weight assignment search is to find the weights that allow to minimise the global error function.

Let $f_{obj}(sol)$ be the current objective function that evaluates the quality of a solution $sol$.

Let $c_{sol1,sol2}$ be a comparison between two solutions, $sol_1$ and $sol_2$.

Let $p_c$ be the user preference for the comparison $c$. $p_c$ can be either $\{sol_1\}$ (the user prefers the solution $sol_1$), $\{sol_2\}$ (the user prefers the solution $sol_2$) or $\{sol_1, sol_2\}$ (the two solutions have the same quality for the user).

We define the function $comp(c, f_{obj}, p_c)$ that determines for a comparison $c$ if the user preference $p_c$ is compatible with the objective function $f_{obj}$, i.e. if the preference formulated by the user is consistent with the quality order obtained by applying the objective function on the solutions. If the user preference $p_c$ is compatible with the objective function $f_{obj}$ for the comparison $c$, $comp(c, f_{obj}, p_c)$ is equal to 0; otherwise it is equal to 1.

$$comp(c, f_{obj}, p_c) = \begin{cases} 0 & if \begin{cases} p_c = \{sol_1, sol_2\} & and \quad f_{obj}(sol_1) = f_{obj}(sol_2) \\ or \quad p_c = \{sol_1\} & and \quad f_{obj}(sol_1) > f_{obj}(sol_2) \\ or \quad p_c = \{sol_2\} & and \quad f_{obj}(sol_2) > f_{obj}(sol_1) \end{cases} \\ 1 & otherwise \end{cases}$$

We define the function $error(c, f_{obj}, p_c)$ that returns the error value for a comparison $c$. This function is defined as:

$$error(c, f_{obj}, p_c) = \begin{cases} 0 & \text{if } comp(c, f_{obj}, p_c) = 0 \\ val_{error} + |f_{obj}(sol_1) - f_{obj}(sol_2)| & \text{if } comp(c, f_{obj}, p_c) = 1 \end{cases}$$

In this function, we integrated a parameter $val_{error}$ that represents the minimum importance of an error whatever the values of the objective function for the two solutions are. The higher the value of this parameter, the more important it will be to minimise the number of incompatible comparisons.

Finally, the global error function proposed corresponds to the mean error obtained with the objective function $f_{obj}$ on the comparison sample $Comp$:

$$Error(f_{obj}, Comp) = \frac{1}{|Comp|} \times \sum_{c \in Comp} error(c, f_{obj}, p_c)$$

The aims of the weight assignment step is to find a weight assignment that minimises $Error(f_{obj}, Comp)$. The size of the search space will be most of time too high to carry out a complete search. Thus, it will be necessary to proceed by incomplete search. In this context, we propose to use a metaheuristic to find the best weight assignment. In the literature, numerous metaheuristics were introduced [8]-[11]-[15]. In this paper, we propose to use genetic algorithms [12] which are particularly effective when the search space is well-structured as it is in our search problem.

*2) Partitioning of the measure space*

For some user need definition problems, it will not be possible to find a weight assignment compatible with all user preferences. Thus, we propose to partition the measure set space and to define for each partition a regression rule with its own weight assignment.

We propose to base our partitioning method on the utilisation of supervised learning techniques. The goal is to search the parts of the measure space that have a different behaviour in terms of objective function. Thus, we search to detect the parts of the measure space which contain solutions linked to an incompatible comparison.

We built an example set composed of solutions described by its measures values. The conclusion could be either "compatible" if the comparison which contains the solution is compatible with the objective function or "incompatible" if it is not. Then, a supervised learning algorithm is used to partition the measure space. We remind that we proposed to express the partition in the form of rules. Thus, it is necessary to use a supervised learning algorithm that allows to build a predictive model expressed by rules. Different algorithms could be used for this partitioning problem such as RIDOR [10] or C4.5 algorithm [16]. In this paper, we propose to use the effective and well-established RIPPER algorithm [7].

Figure 4 presents an example of partitioning for a measure set composed of two measures.

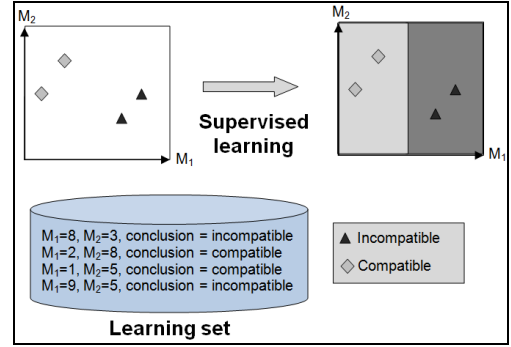

Fig. 4. Partitioning method

Once the partitioning is carried out, the user need definition module performs a new search of the best weights assignment. All partitions are considered at the same time for this search. If the weights assignment found is better (in terms of minimisation of the global error value) than the assignment obtained before the partitioning, the new objective function is kept. Otherwise, the module keeps the previously obtained objective function.

IV. APPLICATION TO CARTOGRAPHIC GENERALISATION

*A. Automatic cartographic generalisation*

We propose to test our objective function designing approach in the domain of cartographic generalisation. Cartographic generalisation is the process that aims at simplifying geographic data to suit the scale and purpose of a map. The objective of this process is to ensure the readability of the map while keeping the essential information of the initial data. Figure 5 gives an example of cartographic generalisation.

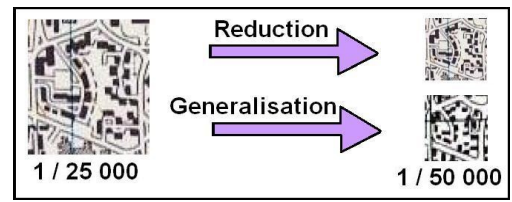

Fig. 5. Cartographic Generalisation

The automation of the generalisation process is an interesting industrial application context which is far from being solved. Moreover, it directly interests the mapping agencies that wish to improve their map production lines. At last, the multiplication of web sites allowing creating one's own map increases the needs of reliable and effective automatic generalisation processes.

One classical approach to automate the generalisation process is to use a local, step-by-step and knowledge-based method [4]: each vector object of the database (representing a building, a road segment, etc.) is transformed by application of a sequence of generalisation algorithms realising atomic transformations. The choice of the applied sequence algorithms is not predetermined but built on the fly for each object according to heuristics and to its characteristics.

## B. The generalisation system

The generalisation system that we use for our experiment is based on the AGENT model [3]-[17]. The AGENT model has been further described in [18].

In this model, geographic objects (roads, buildings, etc) are modelled as agents. Geographic agents manage their own generalisation, choosing and applying generalisation operations to themselves. Each state of the agent represents the geometric state of the considered geographic objects.

During its generalisation process, each agent is guided by a set of constraints that represents the specifications of the desired cartographic product. An example of constraint is, for a building agent, to be big enough to be readable. Each constraint has a satisfaction level between 0 (constraint not satisfy at all) and 100 (constraint perfectly satisfy). For each state, the agent computes its own satisfaction according to the values of its constraint satisfaction.

To satisfy its constraints as well as possible, a geographical agent carries out a cycle of actions during which it tests different actions in order to reach a perfect state (where all of its constraints are perfectly satisfied) or at least the best possible state. The action cycle results in an informed exploration of a state tree. Each state represents the geometric state of the considered geographic objects. Figure 6 gives an example of a state tree obtained with the generalisation system.

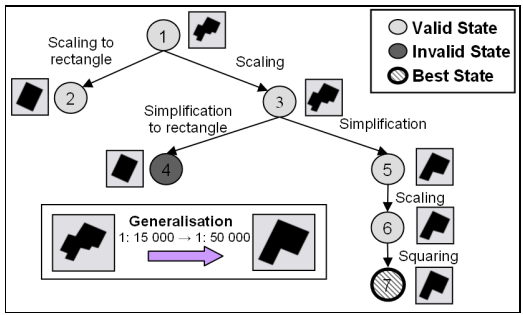

Fig. 6. Example of a state tree for the generalisation of a building

## C. Difficulties of the agent satisfaction function definition

The AGENT model has been the core of numerous research works and is used for map production in several mapping agencies. However, the question of the evaluation of the state of an agent is still asked. The function usually used is a mean of the constraint satisfaction weighted by their importance (which is often an integer ranged between 1 and 10). The problem of this function is to give satisfaction values too homogenous. More over, it does not allow to take into account discontinuities in the satisfaction function. At last, the definition of the importance values is often complex and fastidious when more than five constraints are in stake [2]. Thus, having an approach like the one described in this paper allowing to design the agent satisfaction function is particularly interesting in the context of the AGENT model.

## D. Implementation of our approach for the AGENT model

We experiment our approach on an implementation of our user need definition module in Java, using GéOxygene [2] for geographical data transformation, and *WEKA* [22] for the partitioning part using RIPPER algorithm.

Figure 7 presents our implemented interface. On the top panel, the initial state for a building is presented to the user, with, under, the two possible solutions. The user gives its preference for this sample.

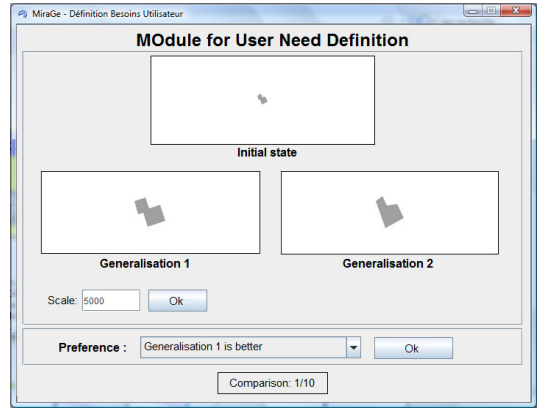

Fig. 7. Implemented graphic interface

## E. Case study

### 1) Setting of the case study

We propose to apply our user need definition approach for the learning of the satisfaction function of the generalisation of *building* agents for 1:25000 scaled maps.

We defined six constraints for the *building* agents:
- *Size constraint*: the building shape should be big enough. Let $S_{sz}$ be the value of this constraint satisfaction.
- *Granularity constraint*: the building shape should not contain too small details. Let $S_{gr}$ be the value of this constraint satisfaction.
- *Squareness constraint*: the angles of the building that are nearly square should be square. Let $S_{sq}$ be the value of this constraint satisfaction.
- *Convexity constraint*: the convexity of the building should be preserved. Let $S_{cv}$ be the value of this constraint satisfaction
- *Elongation constraint*: the elongation of the building should be preserved. Let $S_{el}$ be the value of this constraint satisfaction
- *Orientation constraint*: the orientation of the building should be preserved. Let $S_{or}$ be the value of this constraint satisfaction

### 2) Experiment protocol

50 comparisons (the *learning* set) were presented to a generalisation expert to learn an objective function. Then, we tested the learnt objective function on 50 new comparisons (the *test* set) which were selected in a new area and for which the expert expressed its preferences.

The value used for $val_{error}$ (cf. Section III.D.1) is 40. This value is high enough to limit the number of incompatible comparisons and, at the same time, not too high in order to take into account the difference of values of the objective function value in case of errors. Thus, in our application context, the value of the global error is ranged between 0 and 140.

*3) Results*

The learnt objective function (with *S*, the satisfaction of the building agent) is the following:

$$\begin{cases} if (S_{cv} < 83) \Rightarrow & S = \frac{1}{28}(6 \times S_{cv} + 2 \times S_{el} + 9 \times S_{sq} + 2 \times S_{gr} + 2 \times S_{or} + 7 \times S_{sz}) \\ if (83 \leq S_{cv} < 93) \Rightarrow & S = \frac{1}{28}(7 \times S_{cv} + 7 \times S_{el} + 1 \times S_{sq} + 7 \times S_{gr} + 6 \times S_{sz}) \\ if (S_{cv} > 93) \Rightarrow & S = \frac{1}{30}(9 \times S_{cv} + 9 \times S_{el} + 3 \times S_{sq} + 2 \times S_{gr} + 7 \times S_{sz}) \end{cases}$$

Table 1 presents the results obtained on the two comparison sets. The learnt objective function allowed to get, for both comparison sets, a global error value lower than 5 and a number of incompatible comparisons equals to 5.

|  | *Nb of incompatible comparisons* | *Global error* |
|---|---|---|
| **Learning set** | 5 | 4.25 |
| **Test set** | 5 | 4.63 |

Table 1. Results of the learnt objective function on the learning set and on the test set.

These results show that our approach allowed to learn a pertinent objective function. Indeed, the results obtained by the learnt function are both good on the learning set and on the test set. For both comparison sets, the global error value is very low and only 5 of the 50 comparisons are incompatible. Among these incompatible comparisons, several can be explained by the lack of pertinent measures used to describe the generalisation results. Indeed, the *Measure consistency analysis* comparison choice strategy allowed us to detect that, for some comparisons, two states can be identical in terms of constraint satisfactions but different in terms of generalisation quality.

## V. CONCLUSION

In this paper, we presented an approach dedicated to the definition of user needs. Thus, we proposed an approach based on a man-machine dialogue aiming at defining an objective function representing the user expectations toward an optimisation system. An experiment, carried out in the domain of cartographic generalisation, showed that our approach can help users to formalise their needs and can allow to detect lacks of pertinent measures.

Our approach is based on the utilisation of comparison choosing strategies. In this paper, we defined four different strategies. Other strategies, more complex, could be proposed, such as strategies that take more into account the preferences initially formalised by the user.

Concerning the exploration part as well as the partitioning part, we just tested one search algorithm and one supervised learning algorithm. An interesting study could be to test others algorithms and to compare the results with the ones obtained.

A last perspective could be to pass from an acquisition problem to a revision problem. Indeed, it could be interesting to take into account an initial objective function and to refine it rather than learning a new one from scratch.